# On the Complexity of Solving Markov Decision Problems

Michael L. Littman, Thomas L. Dean, Leslie Pack Kaelbling
Department of Computer Science
Brown University
115 Waterman Street
Providence, RI 02912, USA
Email: {mlittman,tld,lpk}@cs.brown.edu

## Abstract

Markov decision problems (MDPs) provide the foundations for a number of problems of interest to AI researchers studying automated planning and reinforcement learning. In this paper, we summarize results regarding the complexity of solving MDPs and the running time of MDP solution algorithms. We argue that, although MDPs can be solved efficiently in theory, more study is needed to reveal practical algorithms for solving large problems quickly. To encourage future research, we sketch some alternative methods of analysis that rely on the structure of MDPs.

## 1 INTRODUCTION

A *Markov decision process* is a controlled stochastic process satisfying the Markov property with costs assigned to state transitions. A *Markov decision problem* is a Markov decision process together with a performance criterion. A solution to a Markov decision problem is a *policy*, mapping states to actions, that (perhaps stochastically) determines state transitions to minimize the cost according to the performance criterion. Markov decision problems (MDPs) provide the theoretical foundations for decision-theoretic planning, reinforcement learning, and other sequential decision-making tasks of interest to researchers and practitioners in artificial intelligence and operations research (Dean et al., 1995b). MDPs employ dynamical models based on well-understood stochastic processes and performance criteria based on established theory in operations research, economics, combinatorial optimization, and the social sciences (Puterman, 1994).

It would seem that MDPs exhibit special structure that might be exploited to expedite their solution. In investment planning, for example, often the initial state is known with certainty (the current price for a stock or commodity) and as a result the set of likely reachable states (future prices) and viable investment strategies in the near-term future is considerably restricted. In general, notions of time, action, and reachability in state space are inherent characteristics of MDPs that might be exploited to produce efficient algorithms for solving them. It is important that we understand the computational issues involved in these sources of structure to get some idea of the prospects for efficient sequential and parallel algorithms for computing both exact and approximate solutions.

This paper summarizes some of what is known (and unknown but worth knowing) about the computational complexity of solving MDPs. Briefly, any MDP can be represented as a linear program (LP) and solved in polynomial time. However, the order of the polynomials is large enough that the theoretically efficient algorithms are not efficient in practice. Of the algorithms specific to solving MDPs, none is known to run in worst-case polynomial time. However, algorithms and analyses to date have made little use of MDP-specific structure, and results in related areas of Monte Carlo estimation and Markov chain theory suggest promising avenues for future research. We begin by describing the basic class of problems.

## 2 MARKOV DECISION PROBLEMS

For our purposes, a *Markov decision process* is a four-tuple $(\Omega_S, \Omega_A, p, c)$, where $\Omega_S$ is the *state space*, $\Omega_A$ is the *action space*, $p$ is the state-transition probability-distribution function, and $c$ is the instantaneous-cost function.

The state-transition function is defined as follows: for all $i, j \in \Omega_S, k \in \Omega_A$,

$$p_{ij}^k = \Pr(S_t = j | S_{t-1} = i, A_t = k)$$

where $S_t$ ($A_t$) is a random variable denoting the state (action) at time $t$. The cost $c_i^k$ is defined to be the cost of taking action $k$ from state $i$.

Let $N = |\Omega_S|$ and $M = |\Omega_A|$. For some of the complexity results, it will be necessary to assume that $p$ and $c$ are encoded using $N \times N \times M$ tables of rational numbers. We let $B$ be the maximum number of bits required to represent any component of $p$ or $c$. In this paper, we restrict our attention to discrete-time processes in which both $\Omega_S$ and $\Omega_A$ are finite.

A Markov decision process describes the dynamics of an agent interacting with a stochastic environment. Given an initial state or distribution over states and a sequence of actions, the Markov decision process describes the subsequent evolution of the system state over a (possibly infinite) sequence of times referred to as the *stages* of the process. This paper focuses on the *infinite-horizon* case, in which the sequence of stages in infinite.

A *policy* $\pi$ is a mapping from states to actions. If the policy is independent of the current stage, it is said to be *stationary*.

A *Markov decision problem* (MDP) is a Markov decision process together with a *performance criterion*. The performance criterion enables us to assign a *total cost* to each state for a given policy. A policy and a Markov decision process, together with an initial state, determine a probability distribution over sequences of state/action pairs called *trajectories*. The performance criterion assigns to each such trajectory a cost (determined in part by the instantaneous-cost function) and the probability-weighted sum of these costs determine the policy's total cost for that state.

A policy $\pi_1$ is said to *dominate* policy $\pi_2$ if, for every state $i \in \Omega_S$ the total cost of performing $\pi_1$ starting in state $i$ is less than or equal to the total cost of performing $\pi_2$ starting in state $i$, and if there is at least one state $j \in \Omega_S$ from which the total cost of performing $\pi_1$ is strictly less than that of $\pi_2$. A fundamental result in the theory of MDPs is that there exists a stationary policy that dominates or has equal total cost to every other policy (Bellman, 1957). Such a policy is termed an *optimal policy* and the total cost it attaches to each state is said to be the optimal total cost for that state. An $\epsilon$-*optimal* solution to a Markov decision problem is a policy whose total cost, for every state, is within $\epsilon$ of the optimal total cost. For the problems we are interested in, the optimal total-cost function (mapping from states to their optimal total costs) is unique but the optimal policy need not be.

We briefly consider three popular performance criteria: *expected cost to target*, *expected discounted cumulative cost*, and *average expected cost per stage*. In the expected cost-to-target criterion, a subset of $\Omega_S$ is designated as a target and the cost assigned to a trajectory is the sum of the instantaneous costs until some state in the target set is reached. In the expected discounted cumulative cost criterion, the cost of a trajectory is the sum over all $t$ of $\gamma^t$ times the instantaneous cost at time $t$, where $0 < \gamma < 1$ is the *discount rate* and $t$ indicates the stage.[1] Under reasonable assumptions (Derman, 1970), the expected cost to target and expected discounted cumulative cost criteria give rise to equivalent computational problems. The average expected cost per stage criterion is attractive because it does not require the introduction of a seemingly arbitrary discount rate, nor the specification of a set of target states. However, it is often a difficult criterion to analyze and work with.

This paper focuses on the expected discounted cumulative cost criterion. To simplify the notation, suppose that the instantaneous costs are dependent only upon the initial state and action so that, for each $i \in \Omega_S$ and $k \in \Omega_A$, $c_i^k = c_{ij}(k)$ for all $j \in \Omega_S$. The expected discounted cumulative cost with respect to a state $i$ for a particular policy $\pi$ and fixed discount $\gamma$ is defined by the following system of equations: for all $i \in \Omega_S$,

$$E_\pi(\Sigma_\gamma | i) = c_i^{\pi(i)} + \gamma \sum_{j \in \Omega_S} p_{ij}^{\pi(i)} E_\pi(\Sigma_\gamma | j). \quad (1)$$

The *optimal total-cost function* $E^*(\Sigma_\gamma | \cdot)$ is defined as

$$E^*(\Sigma_\gamma | i) = \min_\pi E_\pi(\Sigma_\gamma | i), \quad i \in \Omega_S,$$

which can be shown to satisfy the following *optimality equations*: for all $i \in \Omega_S$,

$$E^*(\Sigma_\gamma | i) = \min_{k \in \Omega_A} \left[ c_i^k + \gamma \sum_{j \in \Omega_S} p_{ij}^k E^*(\Sigma_\gamma | j) \right]. \quad (2)$$

This family of equations, due to Bellman (1957), is the basis for several practical algorithms for solving MDPs. There is a policy, $\pi^*$, called the *optimal policy*, which achieves the optimal total-cost function. It can be found from the optimal total-cost function as follows: for all $i \in \Omega_S$,

$$\pi^*(i) = \arg \min_{k \in \Omega_A} \left[ c_i^k + \gamma \sum_{j \in \Omega_S} p_{ij}^k E^*(\Sigma_\gamma | j) \right]. \quad (3)$$

## 3  GENERAL COMPLEXITY RESULTS

There is no known algorithm that can solve general MDPs in a number of arithmetic operations polynomial in $N$ and $M$. Such an algorithm would be called *strongly polynomial*. Using linear programming, however, the problem can be solved in a number of arithmetic operations polynomial in $N$, $M$, and $B$.

Papadimitriou and Tsitsiklis (1987) analyzed the computational complexity of MDPs. They showed that, under any of the three cost criteria mentioned earlier, the problem is P-complete. This means that, although it is solvable in polynomial time, if an efficient parallel algorithm were available, then all problems in P would be solvable efficiently in parallel (an outcome considered unlikely by researchers in the field). Since the linear programming problem is also P-complete, this result implies that in terms of parallelizability, MDPs and LPs are equivalent: a fast parallel algorithm for solving one would yield a fast parallel algorithm for solving the other. It is not known whether the two problems are equivalent with respect to strong polynomiality: although it is clear that a strongly polynomial algorithm for solving linear programs would yield one for MDPs, the inverse is still open.

---

[1] When $\gamma$ is considered part of the input, it is assumed to be encodable in $B$ bits.

Papadimitriou and Tsitsiklis also show that for MDPs with deterministic transition functions (the components of $p$ are all 0's and 1's), optimal total-cost functions can be found efficiently in parallel for all three cost criteria (*i.e.*, the problem is in NC). Further, the algorithms they give are strongly polynomial. This suggests that the stochastic nature of some MDPs has important consequences for complexity and that not all MDPs are equally difficult to solve.

## 4   ALGORITHMS AND ANALYSIS

This section describes the basic algorithms used to solve MDPs and analyzes their running times.

### 4.1   LINEAR PROGRAMMING

The problem of computing an optimal total-cost function for an infinite-horizon discounted MDP can be formulated as a linear program (LP) (D'Epenoux, 1963). Linear programming is a very general technique and does not appear to take advantage of the special structure of MDPs. Nonetheless, this reduction is currently the only proof that MDPs are solvable in polynomial time.

The *primal* linear program involves maximizing the sum

$$\sum_{j \in \Omega_S} v_j$$

subject to the constraints

$$v_i \leq c_i^k + \gamma \sum_{j \in \Omega_S} p_{ij}^k v_j, \tag{4}$$

for all $i \in \Omega_S, k \in \Omega_A$, where $v_i$ for $i \in \Omega_S$ are the variables that we are solving for and which, for an optimal solution to the linear program, determine the optimal total-cost function for the original MDP. The intuition here is that, for each state $i$, the optimal total cost from $i$ is no greater than what would be achieved by first taking action $k$, for each $k \in \Omega_A$. The maximization insists that we choose the greatest lower bound for each of the $v_i$ variables.

It is also of interest to consider the *dual* of the above program which involves minimizing the sum

$$\sum_{i \in \Omega_S} \sum_{k \in \Omega_A} x_i^k c_i^k$$

subject to the constraints

$$\sum_{k \in \Omega_A} x_j^k = 1 + \gamma \sum_{i \in \Omega_S} \sum_{k \in \Omega_A} p_{ij}^k x_i^k, \tag{5}$$

for all $j \in \Omega_S$. The $x_j^k$ variables can be thought of as indicating the amount of "policy flow" through state $j$ that exits via action $k$. Under this interpretation, the constraints are flow conservation constraints that say that the total flow exiting state $j$ is equal to the flow beginning at state $j$ (always 1) plus the flow entering state $j$ via all possible combinations of states and actions weighted by their probability. The objective, then, is to minimize the cost of the flow.

If $\{x_i^k\}$ is a feasible solution to the dual, then $\sum_{i \in \Omega_S} \sum_{k \in \Omega_A} x_i^k c_i^k$ can be interpreted as the total cost of the stationary *stochastic* policy that chooses action $k$ in state $i$ with probability

$$x_i^k / \sum_{k \in \Omega_A} x_i^k.$$

This solution can be converted into a deterministic optimal policy as follows:

$$\pi^*(i) = \arg\max_{k \in \Omega_A} x_i^k.$$

The primal LP as $NM$ constraints and $N$ variables and the dual $N$ constraints and $NM$ variables. In both LPs, the coefficients have a number of bits polynomial in $B$. There are algorithms for solving rational LPs that take time polynomial in the number of variables and constraints as well as the number of bits used to represent the coefficients (Karmarkar, 1984; Khachian, 1979). Thus, MDPs can be solved in time polynomial in $N$, $M$, and $B$. A drawback of the existing polynomial time algorithms is that they run extremely slowly in practice and so are rarely used.

The most popular (and practical) methods for solving linear programs are variations of Dantzig's (1963) simplex method. The simplex method works by choosing subsets of the constraints to satisfy with equality and solving the resulting linear equations for the values of the variables. The algorithm proceeds by iteratively swapping constraints in and out of the selected subset, continually improving the value of the objective function. When no swaps can be made to improve the objective function, the optimal solution has been found. Simplex methods differ as to their choice of *pivot rule*, the rule for choosing which constraints to swap in and out at each iteration.

Although simplex methods seem to perform well in practice, Klee and Minty (1972) showed that one of Dantzig's choices of pivoting rule could lead the simplex algorithm to take an exponential number of iterations on some problems. Since then, other pivoting rules have been suggested and almost all have been shown to result in exponential running times in the worst case. None has been shown to result in a polynomial-time implementation of simplex. Note that these results may not apply directly to the use of linear programming to solve MDPs since the set of linear programs resulting from MDPs might not include the counterexample linear programs. This is an open issue.

There are two ways to consider speeding up the solutions of MDPs: finding improved methods for solving LPs or using solution methods that are specific to MDPs. While progress has been made on speeding up linear programming algorithms (such as a subexponential simplex algorithm which uses a randomized pivoting rule (Bland, 1977; Kalai, 1992)), MDP-specific algorithms hold more promise for efficient solution. We address such algorithms, specifically policy iteration and value iteration, in the following sections.

### 4.2 POLICY ITERATION

The most widely used algorithms for solving MDPs are iterative methods. One of the best known of these algorithms is due to Howard (1960) and is known as *policy iteration*. Policy iteration alternates between a *value determination* phase, in which the current policy is evaluated, and a *policy improvement* phase, in which an attempt is made to improve the current policy.

Policy improvement can be performed in $O(MN^2)$ arithmetic operations (steps), and value determination in $O(N^3)$ steps by solving a system of linear equations.[2] The total running time, therefore, is polynomial if and only if the number of iterations required to find an optimal or $\epsilon$-optimal policy is polynomial. This question is addressed later in the section.

The basic policy iteration algorithm works as follows:

1. Let $\pi'$ be a deterministic stationary policy.

2. Loop

   (a) Set $\pi$ to be $\pi'$.
   (b) Determine, for all $i \in \Omega_S$, $E_\pi(\Sigma_\gamma|i)$ by solving the set of $N$ equations in $N$ unknowns described by Equation 1.
   (c) For each $i \in \Omega_S$, if there exists some $k \in \Omega_A$ such that
   $$\left[c_i^k + \gamma \sum_{j \in \Omega_S} p_{ij}^k E_\pi(\Sigma_\gamma|j)\right] < E_\pi(\Sigma_\gamma|i),$$
   then set $\pi'(i)$ to be $k$, otherwise set $\pi'(i)$ to be $\pi(i)$.
   (d) Repeat loop if $\pi \neq \pi'$

3. Return $\pi$.

Step 2b is the value determination phase and Step 2c is the policy improvement phase.

Since there are only $M^N$ distinct policies, and each new policy dominates the previous one (Puterman, 1994), it is obvious that policy iteration terminates in at most an exponential number of steps. We are interested in finding a polynomial upper bound or in showing that no such upper bound exists (*i.e.*, that the number of iterations can be exponential in the worst case).

---
[2]In theory, value determination can probably be done somewhat faster, since it primarily requires inverting a $N \times N$ matrix, which can be done in $O(N^{2.376})$ steps (Coppersmith and Winograd, 1987).

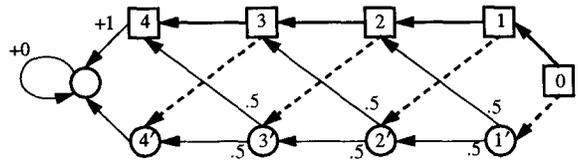

Figure 1: Simple policy iteration requires an exponential number of iterations to generate an optimal solution to the family of MDPs illustrated here (after (Melekopoglou and Condon, 1990)).

While direct analyses of policy iteration have been scarce, several researchers have examined a sequential-improvement variant of policy iteration, in which the current policy is improved for at most one state in Step 2c. A detailed analogy can be constructed between the choice of state to update in sequential-improvement policy iteration and the choice of pivot rule in simplex. Denardo (1982) shows that the feasible bases for the primal LP (Equation 4) are in one-to-one correspondence with the stationary deterministic policies.

As with simplex, examples have been constructed to make sequential-improvement policy iteration require exponentially many iterations. Melekopoglou and Condon (1990) examine the problem of solving expected cost-to-target MDPs using several variations on the sequential improvement policy iteration algorithm. In a version they call *simple policy iteration*, every state is labeled with a unique index and, at each iteration, the policy is updated at the state with the smallest index of those at which the policy can be improved. They show that the family of counterexamples suggested by Figure 1, from a particular starting policy, takes an exponential number of iterations to solve using simple policy iteration.

A counterexample can be constructed for each even number, $N$ ($N = 10$ in the figure). The states are divided into three classes: decision states (labeled 0 through $N/2 - 1$), random states (labeled $1'$ through $(N/2 - 1)'$), and an absorbing state. From each decision state $i$, there are two actions available: action 0 (heavy solid lines) results in a transition to decision state $i + 1$ and action 1 (dashed lines) results in a transition to random state $(i + 1)'$. From random state $i'$, there is no choice of action and instead a random transition with probability $1/2$ of reaching random state $(i + 1)'$ and probability $1/2$ of reaching decision state $i + 1$ takes place. Actions from decision state $N/2 - 1$ and random state $N/2 - 1$ both result in a transition to the absorbing state. This transition has a cost of $+1$ in the case of decision state $N/2 - 1$ and all other transitions have zero cost.

The initial policy is $\pi_0(i) = 0$, so every decision state $i$

selects the action which takes it to decision state $i+1$. In the optimal policy, $\pi^*(i) = 0$ for $i \neq N/2 - 2$ and $\pi^*(N/2 - 2) = 1$. Although these two policies are highly similar, Melekopoglou and Condon show that simple policy iteration steps through $2^{N/2-2}$ policies before arriving at the optimal policy. We remark that although this example was constructed with the expected cost-to-target criterion in mind, it also holds for the discounted cumulative cost criterion regardless of discount rate.

When the policy is improved at all states in parallel, policy iteration no longer has a direct simplex analogue. It is an open question whether this can lead to exponential running time in the worst case or whether the resulting algorithm is guaranteed to converge in polynomial time. However, we can show that this more popular version of policy iteration is strictly more efficient than the simple policy iteration algorithm mentioned above.

Let $\pi_n$ be the policy found after $n$ iterations of policy iteration. Let $E_{\pi_n}(\Sigma_\gamma|i)$ be the total-cost function associated with $\pi_n$. Let $E^n(\Sigma_\gamma|i)$ be the total-cost function found by value iteration (Section 4.3) starting with $E_{\pi_0}(\Sigma_\gamma|i)$ as an initial total-cost function. Puterman (1994) (Theorem 6.4.6) shows that $E_{\pi_n}(\Sigma_\gamma|i)$ always dominates or is equal to $E^n(\Sigma_\gamma|i)$ and therefore that policy iteration converges no more slowly than value iteration for discounted infinite-horizon MDPs. When combined with a result by Tseng (1990) (described in more detail in the next section) which bounds the time needed for value iteration to find an optimal policy, this shows that policy iteration takes polynomial time, for a fixed discount rate. Furthermore, if the discount rate is included as part of the input as a rational number with the denominator written in unary, policy iteration takes polynomial time. This makes policy iteration a *pseudo-polynomial-time* algorithm.

Thus, whereas policy iteration runs in polynomial time for a fixed discount rate, simple policy iteration can take exponential time, regardless of discount rate. This novel observation stands in contrast to a comment by Denardo (1982). He argues that block pivoting in simplex achieves the same goal as parallel policy improvement in policy iteration and therefore that one should prefer commercial implementations of simplex to home-grown implementations of policy iteration. His argument is based on the misconception that one step of policy improvement on $N$ states is equivalent in power to $N$ iterations of simple policy iteration. In fact, one policy improvement step on $N$ states is more like $2^N$ iterations of simple policy iteration, in the worst case. Thus, policy iteration has not yet been ruled out as the preferred solution method for MDPs. More empirical study is needed.

### 4.3 VALUE ITERATION

Bellman (1957) devised a successive approximation algorithm for MDPs called *value iteration* which works by computing the optimal total-cost function assuming first a one-stage finite horizon, then a two-stage finite horizon, and so on. The total-cost functions so computed are guaranteed to converge in the limit to the optimal total-cost function. In addition, the policy associated with the successive total-cost functions will converge to the optimal policy in a finite number of iterations (Bertsekas, 1987), and in practice this convergence can be quite rapid.

The basic value-iteration algorithm is described as follows:

1. For each $i \in \Omega_S$, initialize $E^0(\Sigma_\gamma|i)$.
2. Set $n$ to be 1.
3. While $n <$ maximum number of iterations,
   (a) For each $i \in \Omega_S$ and $k \in \Omega_A$, let
   $$E^n(\Sigma_\gamma|i,k) = \left[c_i^k + \gamma \sum_{j \in \Omega_S} p_{ij}^k E^{n-1}(\Sigma_\gamma|j)\right].$$
   $$E^n(\Sigma_\gamma|i) = \min_{k \in \Omega_A} E^n(\Sigma_\gamma|i,k)$$
   (b) Set $n$ to $n+1$.
4. For each $i \in \Omega_S$,
   $$\pi(i) = \arg\min_{k \in \Omega_A} E^n(\Sigma_\gamma|i,k).$$
5. Return $\pi$.

The maximum number of iterations is either set in advance or determined automatically using an appropriate *stopping rule*. The *Bellman residual* at step $n$ is defined to be
$$\max_{i \in \Omega_S} \left|E^n(\Sigma_\gamma|i) - E^{n-1}(\Sigma_\gamma|i)\right|.$$
By examining the Bellman residual during value iteration and stopping when it gets below some threshold, $\epsilon' = \epsilon(1-\gamma)/(2\gamma)$, we can guarantee that the resulting policy will be $\epsilon$-optimal (Williams and Baird, 1993).

The running time for each iteration is $O(MN^2)$, thus, once again, the method is polynomial if and only if the total number of iterations required is polynomial. We sketch an analysis of the number of iterations required for convergence to an optimal policy below; more detailed discussion can be found in Tseng's paper.

1. Bound the distance from the initial total-cost function to the optimal total-cost function.
   Let $M = \max_{i \in \Omega_S, k \in \Omega_A} |c_i^k|$, the magnitude of the largest instantaneous cost. The total-cost function for any policy will have components in the range $[-M/(1-\gamma), M/(1-\gamma)]$. Thus any choice of initial total-cost function with components in this range cannot differ from the optimal total-cost function by more than $2M/(1-\gamma)$ at any state.

2. Show that each iteration results in an improvement of a factor of at least $\gamma$ in the distance between the estimated and optimal total-cost functions.

   This is the standard "contraction mapping" result for discounted MDPs (Puterman, 1994).

3. Give an expression for the distance between estimated and optimal total-cost functions after $n$ iterations. Show how this gives a bound on the number of iterations required for an $\epsilon$-optimal policy.

   After $n$ iterations the estimated total-cost function can differ from the optimal total-cost function by no more than $2M\gamma^n/(1-\gamma)$ at any state. Solving for $n$ and using the result relating the Bellman residual to the total cost of the associated policy, we can express the maximum number of iterations needed to find an $\epsilon$-optimal policy as

   $$n^* \leq \frac{B + \log(1/\epsilon) + \log(1/(1-\gamma)) + 1}{1-\gamma} \quad . \quad (6)$$

4. Argue that there is a value for $\epsilon > 0$ for which an $\epsilon$-optimal policy is, in fact, optimal.

   The optimal total-cost function can be expressed as the solution of a linear program with rational components of no more than $B$ bits each (Section 4.1). A standard result in the theory of linear programming is that the solution to such a linear program can be written as rational numbers where each component is represented using a number of bits polynomial in the size of the system and $B$, $B^*$ (Schrijver, 1986).

   This means that if we can find a policy that is $\epsilon = 1/2^{B^*+1}$-optimal, the policy must be optimal.

5. Substitute this value of $\epsilon$ into the bound to get a bound on the number of iterations needed for an exact answer.

   Substituting for $\epsilon$ in Equation 6 reveals that running value iteration for a number of iterations polynomial in $N$, $M$, $B$, and $1/(1-\gamma)$ guarantees an optimal policy.

This analysis shows that, for fixed $\gamma$, value iteration takes polynomial time. It is also useful for constructing an upper bound for policy iteration (see Section 4.2). Although it is not known whether the dependence on $1/(1-\gamma)$ (which can be quite large as $\gamma$ approaches 1) can be dropped for policy iteration, we can show that value iteration can indeed take that long.

Figure 2 illustrates a family of MDPs for which discovering the optimal policy via value iteration takes time proportional to $1/(1-\gamma)\log(1/(1-\gamma))$. It consists of 3 states, labeled 0 through 2. From state 0, action 1 causes a deterministic transition to state 1 and action 2 causes a deterministic transition to state 2. Action 1 has no instantaneous cost but once in state 1, there is a cost of +1 for every time step thereafter. Action 2 has an instantaneous cost of $\gamma^2/(1-\gamma)$ but state 2 is a zero-cost absorbing state.[3] The discounted infinite-horizon cost of choosing action 1 from state 0 is $\gamma/(1-\gamma)$ whereas the total cost for action 2 is $\gamma^2/(1-\gamma)$ (smaller, since $\gamma < 1$). If we initialize value iteration to the zero total-cost function, the estimate of the costs of these two choices are: $\gamma(1-\gamma^n)/(1-\gamma)$ and $\gamma^2/(1-\gamma)$ at iteration $n > 1$. Therefore, value iteration will continue to choose the suboptimal action until iteration $n^*$ where:

$$n^* \geq \frac{\log(1-\gamma)}{\log \gamma} \geq \frac{1}{2}\log\left(\frac{1}{1-\gamma}\right)\frac{1}{(1-\gamma)} \quad .$$

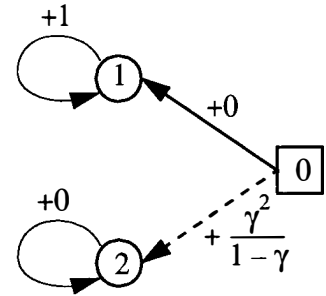

Figure 2: Value iteration requires number of iterations proportional to $1/(1-\gamma)\log(1/(1-\gamma))$ to generate an optimal solution for this family of MDPs.

Thus, in the worst case, value iteration has a running time that grows faster than $1/(1-\gamma)$.

## 5  ALTERNATIVE METHODS OF ANALYSIS

We know that MDPs can be solved in time polynomial in $N$, $M$ and $B$. Unfortunately, the degree of the polynomial is nontrivial and the methods that are guaranteed to achieve such polynomial-time performance do not make any significant use of the structure of MDPs. Furthermore, as with the multi-commodity flow problem (Leighton et al., 1991), the existence of a linear programming solution does not preclude the need for more efficient algorithms, even if it means finding only approximately optimal solutions. This section sketches some directions that could be pursued to find improved algorithms for MDPs.

An in-depth empirical study of existing MDP algorithms might be fruitful. In addition to the solution methods discussed earlier, there are numerous elaborations and hybrids that have been proposed to improve the convergence speed or running time. Puterman and Shin (1978) describe a general method called *modified policy iteration* that includes policy iteration and value iteration as special cases. The structure of modified policy iteration is essentially that of policy iteration where the value determination step is replaced with an approximation that closely resembles value iteration with a fixed policy. Bertsekas (1987) describes variations on value and policy iteration, called *asynchronous* dynamic programming algorithms, that interleave improving policies and estimating the value of policies. These methods resemble techniques used in the reinforcement-learning field where MDPs are solved by performing cost update computations along high probability trajectories. A promising approach from this literature involves a heuristic for dynamically choosing which states to update in value iteration according to how likely such an update would be to improve the estimated total-cost function (Moore and Atkeson, 1993; Peng and Williams, 1993). Before embarking on such a study, we need to compile a suite of benchmark MDPs that reflects interesting classes of problems.

---
[3] Note that these costs can be specified by $B \approx \log(\gamma^2/(1-\gamma)) = O(\log(1/(1-\gamma)))$ bits.

Fast $\epsilon$-approximation algorithms could be very useful in trading off solution accuracy for time. For example, approximation algorithms have been designed for solving linear programs. One is designed for finding $\epsilon$-optimal solutions to a certain class of linear programs which includes the primal linear program given in Section 4.1 (Plotkin et al., 1991). Although this particular scheme is unlikely to yield practical implementations (it is most useful for solving linear programs with exponentially many constraints) the application of approximate linear-programming approaches to MDPs is worth more study.

Probabilistic approximations might also be desirable in some applications, say if we could find an $\epsilon$-optimal solution with probability $1 - \delta$, in some low-order polynomial in $N$, $M$, $1/\epsilon$, $1/\delta$, and $1/(1 - \gamma)$. Fully-polynomial randomized approximation schemes (FPRAS) such as this are generally designed for problems that cannot be computed exactly in polynomial time (*e.g.*, (Dagum and Luby, 1993)), but researchers are now developing iterative algorithms with tight probabilistic performance bounds that provide reliable estimates (*e.g.*, the Dagum *et al.* (1995) optimal stopping rule for Monte Carlo estimation).

Work on FPRAS has identified properties of graphs and Markov chains (*e.g.*, the rapid mixing property for Markov chains used by Jerrum and Sinclair (1988) in approximating the permanent) that may allow us to classify MDPs into easy and hard problems. A related observation is made by Bertsekas (1987) in the context of an algorithm that combines value iteration with a rule for maintaining error bounds. He notes that the convergence of this algorithm is controlled by the discount rate in conjunction with the magnitude of the subdominant eigenvalue of the Markov chain induced by the optimal policy (if it is unique). This value could be used to help characterize hard and easy MDPs.

Some work has already been done to characterize MDPs with respect to their computational properties, including experimental comparisons that illustrate that there are plenty of easy problems mixed in with extraordinarily hard ones (Dean et al., 1995a), and categorization schemes that attempt to relate measurable attributes of MDPs such as the amount of uncertainty in actions to the type of solution method that works best (Kirman, 1994).

One thing not considered by any of the algorithms mentioned above is that, in practice, the initial state is often known. Thus it may be possible to find near-optimal solutions without considering the entire state space (*e.g.*, consider the case in which $\gamma$ is relatively small and it takes many stages to reach more than $\log(N)$ states from the initial state). Dean, Kaelbling, Kirman, and Nicholson (1993) solve MDPs using an algorithm that exploits this property but provide no error bounds on its performance. Barto, Bradtke, and Singh's RTDP (real-time dynamic programming) algorithm (1995) exploits a similar intuition to find an optimal policy without necessarily considering the entire state space.

Structure in the underlying dynamics should allow us to aggregate states and decompose problems into weakly-coupled subproblems, thereby simplifying computation. Aggregation has long been an active topic of research in operations research and optimal control (Schweitzer, 1984). In particular, Bertsekas and Castañon (1989) describe adaptive aggregation techniques that might be very important for large, structured state spaces, and Kushner and Chen (1974) describe how to use Dantzig-Wolfe LP decomposition techniques (1960) for solving large MDPs. More recently, researchers in planning (Boutilier et al., 1995b; Dean and Lin, 1995) and reinforcement learning (Kaelbling, 1993; Moore and Atkeson, 1995) have been exploring aggregation and decomposition techniques for solving large MDPs. What is needed is a clear mathematical characterization of the classes of MDPs for which these techniques guarantee good approximations in low-order polynomial time.

Finally, our preoccupation with computational complexity is not unjustified. Although, in theory, MDPs can be solved in polynomial time in the size of the state space, action space, and bits of precision, this only holds true for so-called *flat* representations of the system dynamics in which the states are explicitly enumerated. Boutilier *et al.* (1995), consider the advantages of *structured* state spaces in which the representation of the dynamics is some log factor of the size of the state space. An efficient algorithm for these MDPs would therefore need to run in time bounded by a polynomial in the *logarithm* of the number of the number of states—a considerably more challenging endeavor.

## 6   SUMMARY AND CONCLUSIONS

In this paper, we focus primarily on the class of MDPs with an expected-discounted-cumulative-cost performance criterion and discount rate $\gamma$. These MDPs can be solved using linear programming in a number of arithmetic operations polynomial in $N$ (the number of states), $M$ (the number of actions), and $B$ (the maximum number of bits required to encode instantaneous costs and state-transition probabilities as rational numbers). There is no known strongly-polynomial algorithm for solving MDPs. The general problem is P-complete and hence equivalent to the problem of solving linear programs with respect to the prospects for exploiting parallelism.

The best known practical algorithms for solving MDPs appear to be dependent on the discount rate $\gamma$. Both value iteration and policy iteration can be shown to perform in polynomial time for fixed $\gamma$, but value iteration can take a number of iterations proportional to $1/(1-\gamma)\log(1/(1-\gamma))$ in the worst case. In addition, a version of policy iteration in which policies are improved one state at a time can be shown to require an exponential number of iterations, regardless of $\gamma$, giving some indication that the standard algorithm for policy iteration is strictly more powerful than this variant. We note that neither value iteration nor policy iteration makes significant use of the structure of the underlying dynamical model.

The fact that the linear programming formulation of MDPs can be solved in polynomial time is not particularly comforting. Existing algorithms for solving LPs with provable polynomial-time performance are impractical for most MDPs. Practical algorithms for solving LPs based on the simplex method appear prone to the same sort of worst-case behavior as policy iteration and value iteration.

We suggest two avenues of attack on MDPs: first, we relax our requirements for performance, and, second, we focus on narrower classes of MDPs that have exploitable structure. The goal is to address problems that are representative of the types of applications and performance expectations found in practice in order to produce theoretical results that are of interest to practitioners.

In conclusion, we find the current complexity results of marginal use to practitioners. We call on theoreticians and practitioners to generate a set of alternative questions whose answers will inform practice and challenge current theory.

### Acknowledgments

Thanks to Justin Boyan, Tony Cassandra, Anne Condon, Paul Dagum, Michael Jordan, Philip Klein, Hsueh-I Lu, Walter Ludwig, Satinder Singh, John Tsitsiklis, and Marty Puterman for pointers and helpful discussion.